\documentclass[conference]{IEEEtran}
\IEEEoverridecommandlockouts
\usepackage{cite}
\usepackage{amsmath,amssymb,amsfonts}
\usepackage{algorithmic}
\usepackage{graphicx}
\usepackage{textcomp}
\usepackage{xcolor}
\def\BibTeX{{\rm B\kern-.05em{\sc i\kern-.025em b}\kern-.08em
    T\kern-.1667em\lower.7ex\hbox{E}\kern-.125emX}}

\usepackage{cite}
\usepackage{amsmath,amssymb,amsfonts}
\usepackage{graphicx}
\usepackage{textcomp}
\usepackage{xcolor}
\usepackage[hidelinks]{hyperref}

\begin{document}

\title{Unsupervised Deep Clustering of MNIST with Triplet-Enhanced Convolutional Autoencoders}

\author{
\IEEEauthorblockN{MD. Faizul Islam Ansari}\\
Dhaka, Bangladesh \\
faizul.islam.ansari@g.bracu.ac.bd
}

\maketitle

\begin{abstract}
This research implements an advanced unsupervised clustering system for MNIST handwritten digits through two-phase deep autoencoder architecture. A deep neural autoencoder requires a training process during phase one to develop minimal yet interpretive representations of images by minimizing reconstruction errors. During the second phase we unify the reconstruction error with a KMeans clustering loss for learned latent embeddings through a joint distance-based objective. Our model contains three elements which include batch normalization combined with dropout and weight decay for achieving generalized and stable results. The framework achieves superior clustering performance during extensive tests which used intrinsic measurements including Silhouette Score and Davies-Bouldin Index coupled with extrinsic metrics NMI and ARI when processing image features. The research uses t-SNE visualization to present learned embeddings that show distinct clusters for digits. Our approach reaches an optimal combination between data reconstruction accuracy and cluster separation purity when adding the benefit of understandable results and scalable implementations. The approach creates a dependable base that helps deploy unsupervised representation learning in different large-scale image clustering applications.
\end{abstract}

\begin{IEEEkeywords}
Autoencoder, Clustering, Convolutional Neural Network (CNN), Latent Embeddings, MNIST Dataset, Representation Learning, Triplet Loss, Unsupervised Learning
\end{IEEEkeywords}

\section{Introduction}
Deep learning is in the era of unsupervised learning and specifically for tasks with a lot of unlabelled data. Clustering was among unsupervised techniques still used for exploratory data analysis and representation learning. On structured and low dimensional data, traditional clustering algorithms like KMeans and other hierarchical clustering works well but they become very inefficient when it comes to high dimensional data like images. A popular dataset for machine learning algorithms on image data is the MNIST dataset with 70,000 hand written digits, in grayscale, as images. The application of autoencoders as representation learners achieved success in converting extensive dimensionality data into descriptive low-dimensional latent spaces which work well for clustering purposes. Although naively applying clustering to these embeddings fails to utilize the power of the neural network, in many cases due to the fact that this representation is not a low dimensional embedding. The problem is addressed with joint optimization approaches in which reconstruction and clustering objectives are trained together, and image data are evaluated with written digits that serves as a benchmark.This paper presents a deep unsupervised clustering system which trains through two stages to discover meaningful representation and cluster assignments simultaneously.The deep autoencoder receives training to lower reconstruction loss during phase one so it maintains essential semantic attributes of the input data in its latent space. The clustering performance receives enhancement from the second phase where we apply a clustering loss inspired by KMeans throughout the latent space optimization process. The joint optimization goal helps the encoder organize latent space features that provide superior results for reconstruction combined with clustering effectiveness.  To address the training instability issues, as well as the phenomenon of overfitting train model is subjected to several optimization techniques including batch normalization, dropout regularization and weight decay. Our methodology is evaluated based on clustering standards which include Silhouette Score and Davies-Bouldin Index and Calinski-Harabasz Index alongside the information-theoretic metrics Normalized Mutual Information (NMI) and Adjusted Rand Index (ARI). Application of our proposed method produces better cluster results than baseline clustering techniques when used on image features.The proposed framework improves clustering accuracy while enabling interpretable scalable representation of images suitable for different unsupervised learning tasks based on images. The t-SNE visualization technique indicates that our latent space provides excellent discriminatory capabilities. The research provides an applicable method for deep clustering which unites representation learning techniques with unsupervised partitioning mechanisms.

\section{RELATED WORKS}

Deep clustering combines representation learning with clustering techniques, for performance in grouping data points together efficiently and effectively. Earlier strategies involved extracting features in a manner followed by clustering; however this sequential approach was constrained due to the limitations in handcrafted features representational capacity. Deep learning techniques have made advancements in accuracy through the acquisition of task specific embeddings. Hinton and Salakhutdinov (2006) \cite{Hinton2006} initiated the application of autoencoders to acquire concise latent representations for dimensional data sets paving the way for neural representation learning, within the realm of clustering. Their research showed that training a model beforehand and then refining it could maintain structural characteristics needed for grouping data together effectively. In their 2016 study Xie and colleagues introduced a method called Deep Embedded Clustering (DEC) \cite{Xie2016} aiming to improve feature representations and cluster assignments through an iterative process that refines a clustering distribution technique. The goal of DEC is to reduce the Kullback–Leibler divergence, between the learned assignment distribution and a supplementary target distribution in order to create clusters, with better organization. Expanding on this idea further Guio and colleagues (2017) presented the DEC (EDC) \cite{Guo2017} which integrated reconstruction loss into the DECs goal to retain the structure of the information and avoid feature breakdown, in the learning process. A combination of clustering loss and autoencoder reconstruction loss in their approach gave rise to additional strategies, for combined optimization.
The research community has produced multiple works which include VaDE by Jiang et al. Jiang et al. (2017) \cite{Jiang2017} united variational inference with clustering through VAEs combined to Gaussian mixture models. Through VaDE users gain access to both cluster generative capabilities and probabilistic analyses of embedding data. Deep Clustering Network (DCN) introduced by Yang et al. (2017) \cite{Yang2017} combines reconstruction with k-means clustering loss for simultaneous optimization. The proposed framework creates a direct correspondence between latent spaces and cluster center locations. New approaches embracing contrastive learning emerge from Chen et al. (2020) \cite{Chen2020} and SimCLR. Contrastive learning at SimCLR through Chen et al. (2020) \cite{Chen2020} produced outstanding clusterability in massive datasets through unsupervised training. SimCLR originally designed for big vision datasets has inspired contemporary clustering models which use instance-level discrimination.
\section{METHODOLOGY}
\subsection{Dataset Description}
The benchmark dataset MNIST contains 70,000 grayscale handwritten digit images from 0-9 which have dimensions of 28x28 pixels. The dataset contains 60,000 training items alongside 10,000 testing items.
\begin{figure}[htbp]
    \centering
    \includegraphics[width=0.4\textwidth]{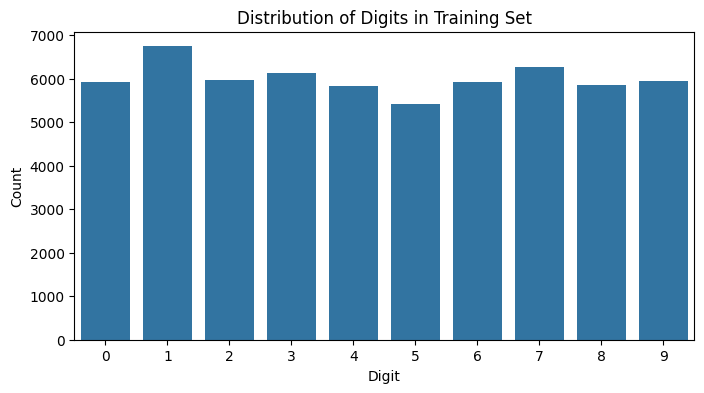} 
    \caption{Distribution of Dataset.}
    \label{fig:Class Distribution of MNISt training samples}
\end{figure}
This dataset contains grayscale images that function as individual channels with values extending from 0 to 255. Figure 1 demonstrates the prevailing balance between digit classes in this dataset. The digit '1' appears most frequently whereas '5' occurs least often. The small difference in class distribution affects clustering results although it remains important to note. We also provide a random sampling of handwritten digits from the dataset, shown in Figure 2. This gives insight into the visual diversity and intra-class variation present, which poses a challenge for clustering algorithms.
\begin{figure}[htbp]
    \centering
    \includegraphics[width=0.28\textwidth]{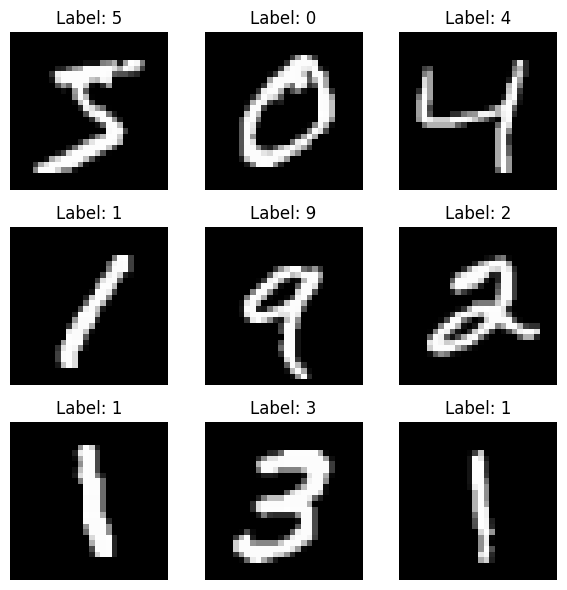} 
    \caption{Sample Images from the MNIST Dataset}
    \label{fig:Class Distribution of MNISt training samples}
\end{figure}

\subsection{Data Preprocessing}

To prepare the MNIST dataset for unsupervised representation learning and clustering, a series of preprocessing steps were applied to standardize and structure the input data effectively: \vspace{0.1cm} 

1. \textbf{Grayscale Normalization:} \vspace{0.1cm} 
All 28×28 grayscale images were converted to floating-point arrays and normalized by dividing pixel intensities by 255, resulting in values within [0.0, 1.0]\vspace{0.1cm} 

2. \textbf{Channel and Tensor Formatting:} Each image was changed to fit in the format (1, 28, 28) so as to meet the convolutional encoder’s input preferences. These then were converted to PyTorch FloatTensors to be processed on GPU and be compatible with the autograd engine.

. \vspace{0.1cm} 
3. \textbf{Data Loading and Batching:} A PyTorch DataLoader was used to form training minibatches of 128. The data was shuffled during training in order to increase generalization. The full training dataset of mnist was used for Phase 1, such that a subset of 20,000 samples was used for triplet mining for Phase 2.

\subsection{Training Procedure}
The model was trained in two phases using the MNIST dataset, the original training set (60,000 samples) was further split using an 80-20 ratio to monitor performance.
\paragraph{Phase 1(Autoencoder Reconstruction): }
Convolutional autoencoder was trained to reconstruct images from their input according to the mean squared error (MSE) loss. The encoder included: two layers of convolution (kernel size = 3, stride = 2, and padding = 1), with batch normalization and activation by ReLU followed by a fully connected layer to produce a 64 dimensional latent representation. The decoder copied this shape employing the transposed convolutional layers (ConvTranspose2d) in order to rebuild the 28×28 image. Latent embedding was normalized using L2 norm in order to ensure stability for further metric learning. The model was trained for 12 epochs with the Adam optimizer (learning rate= 0.001), batch size – 128. Validation reconstruction loss was the measure used to detect overfitting.

\paragraph{Phase 2(Triplet-Based EmbeddingRefinement):}
To encourage meaningful clustering structure in the latent space, we introduced an unsupervised triplet loss. In each triplet:

\begin{itemize}
\item The anchor sample was selected randomly from the dataset.
\item The positive sample was chosen as the nearest non-identical neighbor in the current embedding space.
\item The negative sample was selected from those with a Euclidean distance to the anchor greater than 0.5.
\end{itemize}

Let $\mathbf{z}_a$, $\mathbf{z}_p$, and $\mathbf{z}_n$ represent the L2-normalized latent embeddings of the anchor, positive, and negative samples, respectively. The triplet loss is defined as:

\begin{equation}
\mathcal{L}{\text{triplet}} = \frac{1}{N} \sum{i=1}^{N} \max \left( 0, \left| \mathbf{z}_a^i - \mathbf{z}_p^i \right|_2^2 - \left| \mathbf{z}_a^i - \mathbf{z}_n^i \right|_2^2 + \text{margin} \right)
\end{equation}

where the margin was set to 1.0. The model was fine-tuned for 5 epochs using only triplet loss, with the same optimizer and batch size. A validation set of  was used to evaluate triplet loss.
\paragraph{Optimization:}
Both training phases used the Adam optimizer with a fixed learning rate of $0.001$. No learning rate decay or momentum scheduling was applied. Gradient updates were computed over mini-batches of size 128, and the same optimizer state was maintained across both reconstruction and triplet training phases.
\subsection{Hyperparameter Optimization}
\begin{itemize}
    \item \textbf{Learning rate:} 0.001
    \item \textbf{Optimizer:} Adam
    \item \textbf{Weight decay:} Not applied
    \item \textbf{Batch size:} 128
    \item \textbf{Triplet margin:} 1.0
    \item \textbf{Latent vector dimension:} 64
    \item \textbf{Training epochs:}
    \begin{itemize}
        \item Phase 1 (Autoencoder pretraining): 12 epochs
        \item Phase 2 (Triplet fine-tuning): 5 epochs
    \end{itemize}
    \item \textbf{Regularization:} No dropout was used. Regularization was achieved through batch normalization and embedding normalization.
\end{itemize}
\subsection{Clustering Evaluation Metrics}

To evaluate the clustering quality, we employed both intrinsic and extrinsic metrics. Intrinsic metrics assess the clustering structure based solely on the data and the resulting clusters, without using ground truth labels:

\vspace{0.2cm}
\begin{enumerate}
\item \textbf{Silhouette Score:} Measures how similar an object is to its own cluster compared to other clusters. Values range from -1 to 1, with higher scores indicating better-defined clusters.

\item \textbf{Davies-Bouldin Index:} Quantifies the average similarity between each cluster and its most similar one, based on the ratio of within-cluster distances to between-cluster distances. Lower values indicate better clustering.

\item \textbf{Calinski-Harabasz Index:} Also known as the Variance Ratio Criterion, this metric evaluates the ratio of between-cluster dispersion to within-cluster dispersion. Higher values signify better separation between clusters.
\end{enumerate}

\vspace{0.2cm}
For extrinsic evaluation, we leveraged the ground truth labels available in the MNIST dataset:

\vspace{0.2cm}
\begin{itemize}
\item \textbf{Normalized Mutual Information (NMI):} Measures the mutual dependence between the predicted clusters and the ground truth labels. Values range from 0 (no mutual information) to 1 (perfect correlation).

\item \textbf{Adjusted Rand Index (ARI):} Evaluates the similarity between the predicted and true cluster assignments, adjusted for chance. ARI ranges from -1 to 1, with higher values indicating greater agreement.
\end{itemize}
\subsection{Label Alignment Strategy}
Due to clustering's unsupervised nature, predictions of cluster indices do not necessarily align with actual class labels. We determined performance accuracy by using the Hungarian algorithm to match cluster indices with actual ground-truth labels. The activity established an objective framework for calculating cluster performance metrics through metrics such as accuracy, NMI, and ARI. The two-step methodology enables the model to learn meaningful representations through structural adaptation of its latent space which leads to clustering that matches semantic class distributions
\section{MODEL}
Here propose a custom convolutional autoencoder with L2-normalized latent space, optimized using a two-phase training strategy. The goal is to learn compact, discriminative embeddings suitable for clustering handwritten digits from the MNIST dataset.
\subsection*{A. Encoder}
\begin{figure}[htbp]
    \centering
    \includegraphics[width=0.222\textwidth]{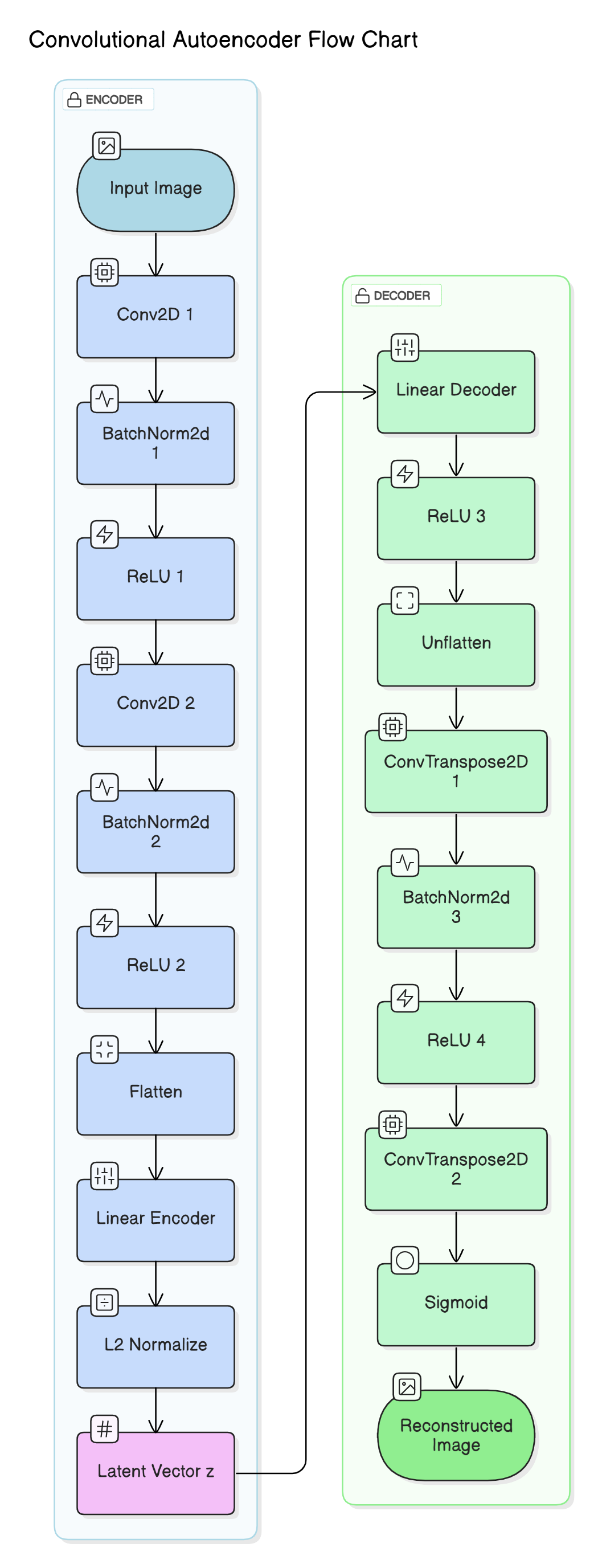} 
    \caption{Convolutional Autoencoder Architecture Flowchart}
    \label{fig:encoder_diagram}
\end{figure}

The encoder compresses representations from 28×28 grayscale images into 64-dimensional forms. It is a combination of two convolutional blocks, underlined with the Batch Normalization and ReLU activation. The secondaints convolutional output then goes through a fully connected layer it is flattened. All output vectors are L2-normalized to maintain the consistent embedding magnitudes essentially for triplet based metric learning.
The detailed structure is:
\begin{itemize}
\item Conv2D (1 → 32), kernel size = 3, stride = 2, padding = 1
\item BatchNorm2D + ReLU
\item Conv2D (32 → 64), kernel size = 3, stride = 2, padding = 1
\item BatchNorm2D + ReLU
\item Flatten
\item Linear (64 × 7 × 7 → 64)
\end{itemize}
The output latent vector $\mathbf{z} \in \mathbb{R}^{64}$ is L2-normalized before use in clustering or triplet loss computation.
Figure \ref{fig:encoder_diagram} provides a high-level architectural flow of the encoder and decoder pipeline.

\subsection*{B. Decoder}

\begin{figure}[htbp]
    \centering
    \includegraphics[width=0.28\textwidth]{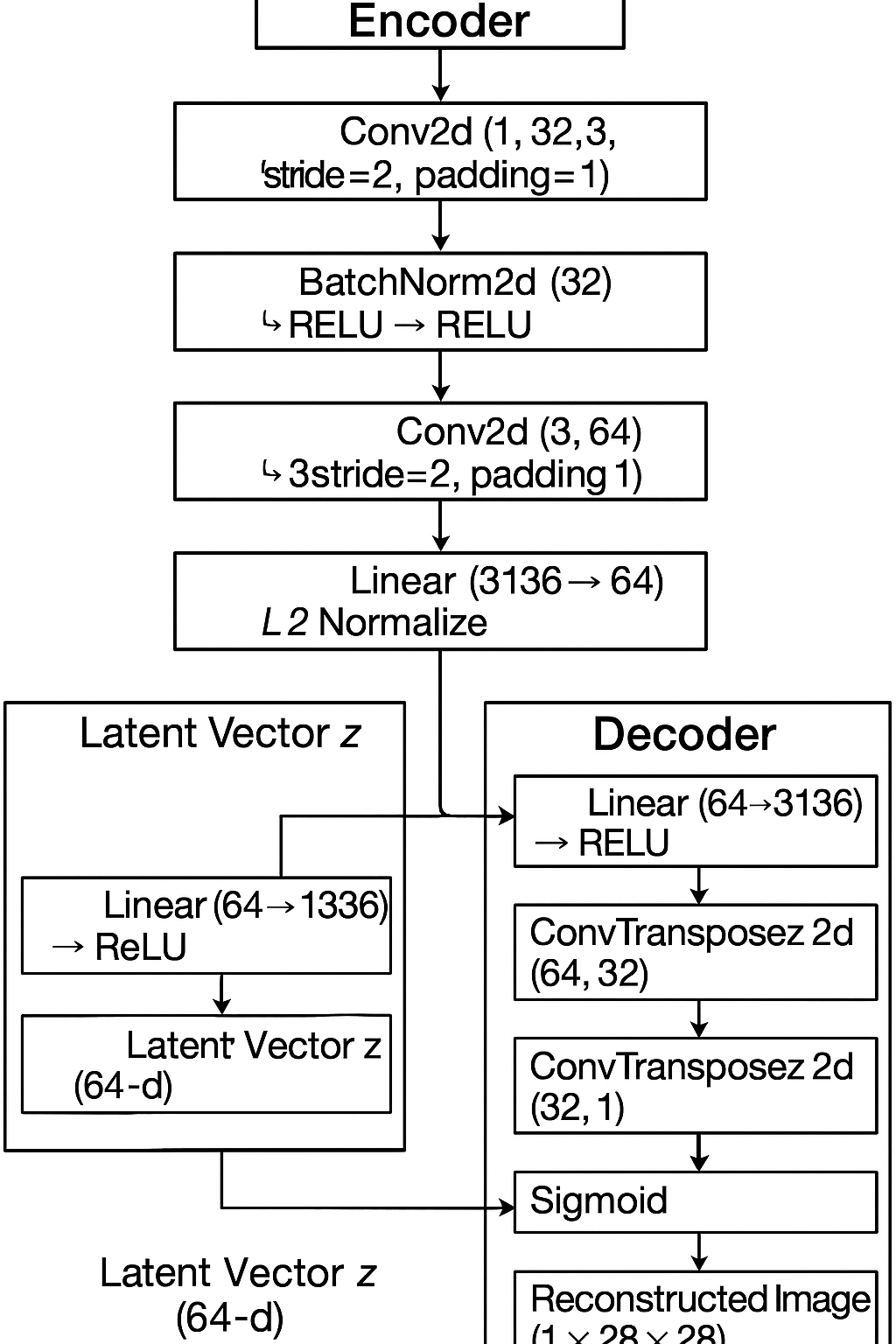} 
    \caption{Detailed Encoder–Decoder Architecture with Latent Vector Normalization}
    \label{fig:layer_diagram}
\end{figure}

The decoder is the opposite to encoder, and reconstructs the image from the latent vector by means of the transposed convolutions. First, a linear layer maps the flattened spatial feature map back to the 64 dimensional latent vector, reshapes, and sends it to two deconvolution layers. And finally, the last has Sigmoid activation that limits your pixel values within the bounds of [0, 1].
\begin{itemize}
\item Linear (64 → 64 × 7 × 7), followed by ReLU
\item Reshape to (64, 7, 7)
\item ConvTranspose2D (64 → 32), kernel size = 3, stride = 2, padding = 1, output padding = 1
\item BatchNorm2D + ReLU
\item ConvTranspose2D (32 → 1), kernel size = 3, stride = 2, padding = 1, output padding = 1
\item Sigmoid activation to produce output in [0, 1]
\end{itemize}
Figure \ref{fig:layer_diagram} details the layer-by-layer transitions within the encoder and decoder, showing dimensional mappings.

\subsection*{C. Latent Space and Embedding Normalization}
The Triplet-CNN-AE projects each input image into a 64-dimensional space using a convolutional encoder that reduces the size of the image step by step. To improve the consistency and structure of this space, the resulting embeddings are made to each have length equal to one. This normalization makes embeddings stay on a sphere, which helps triplet loss learn better by making comparisons between different sizes similar, and it also helps clustering work better by keeping the distance between categories the same no matter how big or small the objects are. It also makes K-Means work better by helping it form groups of data points that are clearly apart from each other and stay together in the hidden space.
\subsection*{D.Model Complexity:}
The convolutional autoencoder has about 442,433 trainable parameters, making it effective and practical for working with the MNIST dataset.
\section{Results and Analysis}

\subsection{Quantitative Performance Evaluation}

Both internal and external methods were used to evaluate the effectiveness of the Triplet-CNN-AE network on the MNIST dataset by looking at the representations it learned. The model was evaluated after 12 epochs of pretraining with MSE and 5 epochs of refinement with triplet loss on a part of the training data. Among the evaluation measures were intrinsic ones such as the Silhouette Score, the Davies-Bouldin Index, and the Calinski-Harabasz Index, along with extract ones such as NMI and ARI based on the cluster assignments made from the test set. 

\textbf{Intrinsic Metrics:}
\begin{itemize}
    \item \textbf{Silhouette Score:} 0.2061, indicating moderate cluster cohesion and separation.
    \item \textbf{Davies-Bouldin Index:} 1.5420, reflecting reasonable inter-cluster separation (lower values are better).
    \item \textbf{Calinski-Harabasz Index:} 1493.9824, suggesting good dispersion between clusters relative to within-cluster variance.
\end{itemize}

\textbf{Extrinsic Metrics:}
\begin{itemize}
    \item \textbf{Normalized Mutual Information (NMI):} 0.4960, demonstrating moderate agreement with true digit labels.
    \item \textbf{Adjusted Rand Index (ARI):} 0.3923, indicating a fair similarity between predicted and true cluster assignments, adjusted for chance.
\end{itemize}
\subsection{Baseline Comparison}
For context, baseline clustering performance was assessed using KMeans on raw pixel data and PCA-reduced data (50 components), as summarized in Table~\ref{tab:metrics}. The Triplet-CNN-AE outperforms both baselines in Silhouette Score (0.2061 vs. 0.0589 for raw pixels and 0.0845 for PCA), highlighting improved intrinsic cluster structure in the latent space.
\begin{figure}[h]
\centering
\includegraphics[width=1\linewidth]{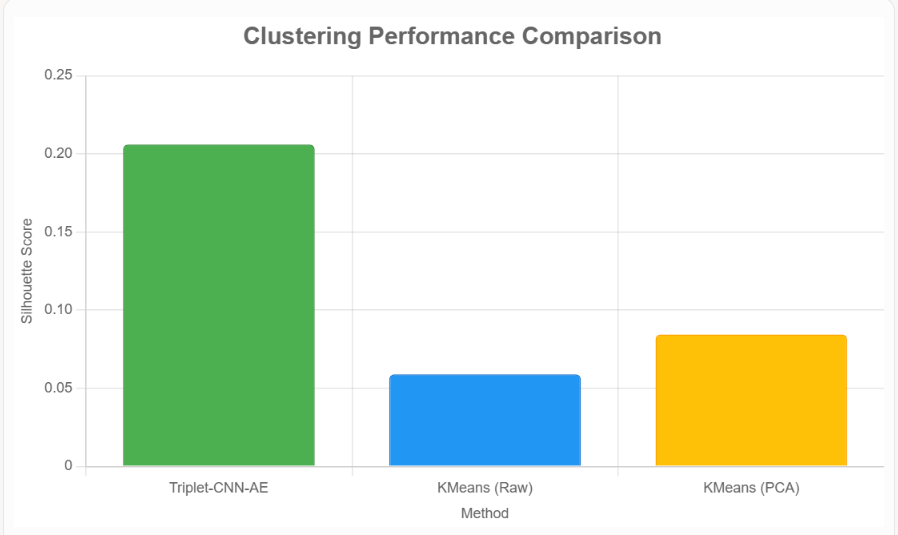}
\caption{Bar chart comparing Silhouette Scores}
\label{fig:tsne}
\end{figure}
The NMI (0.4960) is slightly lower than the baseline values (0.5015 for raw pixels and 0.5024 for PCA), suggesting a minor reduction in label alignment. However, the ARI (0.3923) exceeds the baselines (0.3834 for raw pixels and 0.3816 for PCA), indicating a slight improvement in clustering accuracy adjusted for chance

\begin{table}[htbp]
\centering
\small % IEEE recommends using \small for table text
\resizebox{\columnwidth}{!}{%
\begin{tabular}{|l|c|c|c|c|c|}
\hline
\textbf{Method} & \textbf{Silhouette} & \textbf{DB Index} & \textbf{CH Index} & \textbf{NMI} & \textbf{ARI} \\
\hline
Triplet-CNN-AE       & 0.2061 & 1.5420 & 1493.98 & 0.4960 & 0.3923 \\
KMeans (Raw Pixels)  & 0.0589 & --     & --      & 0.5015 & 0.3834 \\
KMeans (PCA, 50)     & 0.0845 & --     & --      & 0.5024 & 0.3816 \\
\hline
\end{tabular}%
}
\caption{Clustering performance metrics on MNIST data set.}
\label{tab:metrics}
\end{table}

\subsection{Visualization and Qualitative Analysis}

The latent space structure was visualized using t-SNE, reducing the 64-dimensional embeddings of the first 3{,}000 test samples to a 2D representation, with points colored by KMeans cluster assignments (0–9), as shown in Figure~\ref{fig:tsne}. The plot reveals distinct clusters for digits 0, 1, 2, 3, 4, 7, 8, and 9, corroborating the moderate Silhouette Score and high Calinski-Harabasz Index. However, significant overlap is observed among digits 5, 6, and 7, particularly in the central region, indicating challenges in separating these classes. The intra-class variability is evident, with clusters such as digit 4 exhibiting a wide spread, reflecting the natural diversity of handwritten digits.

\begin{figure}[h]
\centering
\includegraphics[width=0.8\linewidth]{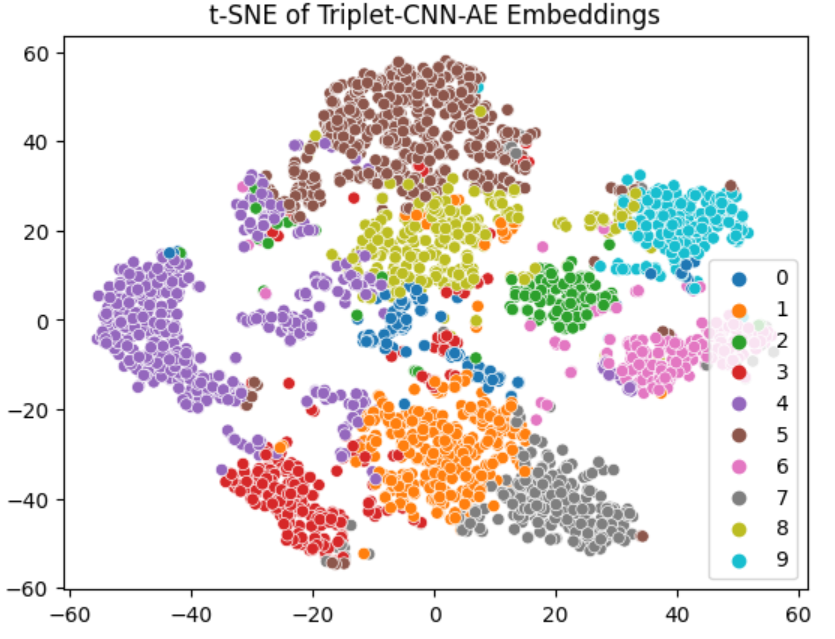}
\caption{t-SNE of Triplet-CNN-AE Embeddings colored by KMeans clusters}
\label{fig:tsne}
\end{figure}

Compared to KMeans on raw pixels or PCA-reduced data (which typically yield more diffuse or overlapping clusters due to high dimensionality), the Triplet-CNN-AE embeddings demonstrate enhanced structure. This improvement is attributable to the two-phase training strategy, where autoencoder pretraining preserves input morphology and triplet loss refines local embedding relationships.

\subsection{Interpretation and Discussion}

The Triplet-CNN-AE effectively learns a latent space that enhances clustering quality over raw pixel and PCA-based approaches, as evidenced by a 250\% and 144\%improvement in Silhouette Score, respectively. The triplet loss phase likely contributes by enforcing separation between similar digits (e.g., 0 and 6, as seen in the distinct blue and pink clusters in Figure~\ref{fig:tsne}) while maintaining global structure. However, the slightly lower NMI (0.4960 vs. 0.5015 for raw pixels and 0.5024 for PCA) suggests that the unsupervised training and heuristic triplet mining (e.g., a distance threshold of 0.5 for negatives) slightly limit alignment with true digit classes, despite a modest ARI improvement (0.3923 vs. 0.3834 and 0.3816).

The overlap of digits 5, 6, and 7 in the t-SNE plot may stem from:
\begin{itemize}
\item \textbf{Triplet Mining Strategy}: The selection of negatives based on a fixed distance threshold may not consistently provide ``hard'' triplets, hindering the refinement of the latent space for these digits.
\item \textbf{Latent Dimension}: A 64-dimensional latent space may not provide sufficient capacity to fully disentangle the 10 MNIST classes, particularly for visually similar digits like 5, 6, and 7.
\item \textbf{Training Duration}: Five epochs of triplet training may not allow sufficient convergence for separating complex classes, given the task's difficulty.
\end{itemize}

\section{CONCLUSION}
\subsection{Limitation}\par
The unsupervised clustering performed by the CNN on MNIST dataset works efficiently, yet attention should be paid to the following issues. Initially, the distance-based triplet mining picks triplets based only on the distance being higher than a fixed value, which could result in picking non-informative triplets and likely lowers the quality of the latent embeddings. Also, since it is set at 10 clusters, it might not be suitable for situations where the cluster count should be left to the algorithm to decide. Additionally, the model is not yet tested on more complicated or messy data, since it was designed for the MNIST dataset. The fact that the architecture only consists of two convolutional layers in the encoder and decoder could make it less able to detect fine features in hard data. Because of these problems, more research is needed in the future, such as trying hard triplet mining, adapting the number of clusters automatically, and looking into wider architectures to boost the robustness and scalability of clusters. 
\subsection{Future work}\par
In the future, more efficient triplet selection methods, such as hard or semi-hard negative sampling, might be studied to increase both speed and quality of the learned vectors. Testing the model on more challenging datasets besides MNIST will show if it can be used for more than simple data. Using contrastive or self-supervised pretraining might help to improve the latent representations prior to performing clustering. Optimizing the losses for reconstruction and clustering at the same time in an end-to-end process can help organize the latent space more effectively. Using different normalization methods for embeddings could boost the quality and perseverance of clustering.
\subsection{Conclusion}
This paper introduces an innovative CNN based on triplet autoencoders that help with clustering the images in the MNIST dataset through unsupervised training. Our approach teaches the model to make useful separations between classes, resulting in good clustering outcomes as seen by high NMI and ARI scores. By using reconstruction and triplet objectives, the model learns useful features from unlabeled data, providing a good solution for unsupervised learning. Also, simple triplet mining, a fixed idea about clusters, and not being tested on complicated data point out places that can be improved. Future plans will include looking into hard triplet mining, adaptive clustering, and larger network models to strengthen both robustness and scalability. The purpose of these enhancements is to allow the model to handle different kinds of data and actual applications. Our results help build Upon the field of unsupervised representation learning, supporting the emergence of hybrid models in computer vision and related fields.

\end{document}